\documentclass[11pt]{article}
\usepackage[a4paper,margin=2.5cm]{geometry}
\usepackage[T1]{fontenc}

\usepackage{mathptmx}
\usepackage{microtype}
\usepackage{tabularx}
\usepackage{amsmath}
\usepackage{booktabs}
\usepackage{graphicx}
\usepackage{xcolor}
\usepackage{enumitem}
\usepackage[hidelinks]{hyperref}
\usepackage{caption}
\setlist{nosep,leftmargin=*}

\title{When More Memory Fails: Hidden Feasibility Holes\\in Dynamic Tensor Rematerialization\\[0.4em]
\large A simulator study}
\author{Mahesh Reddy Pagadala\\Aston University, Birmingham, UK}
\date{October 2026}

\begin{document}
\maketitle

\begin{abstract}
Dynamic Tensor Rematerialization (DTR) trains neural networks under a memory budget by
evicting tensors and recomputing them on demand. A natural expectation is that a budget
large enough to complete training stays sufficient when it is increased. Using DTR's
reference simulator, simrd, to replay the public DTR training traces, we show that this
expectation fails. We call the exceptions \emph{feasibility holes}: memory budgets at which
an eviction policy runs out of memory although both a smaller and a larger budget
complete, a form of non-monotone feasibility. Under a written sweep protocol with dated
amendments, a 0.01-step budget grid shows no such hole for DTR on any of six confirmatory
traces, although unresolved timeouts prevent it from establishing monotonicity. Finer sampling finds holes on InceptionV4 (three regions) and U-Net (two regions),
in addition to one on the ResNet-32 development trace. Every hole is reproduced by three
repeats and by unmodified simrd. For one InceptionV4 pair (0.2343 versus 0.235 of peak
memory) we trace the failure to individual allocations. The larger budget avoids one
eviction at an operator where the smaller budget was 173{,}479 bytes short. The extra data
it carries forward forces a later eviction of a tensor needed three operators afterwards,
and rebuilding that tensor triggers a recomputation cascade that cannot fit. Keeping that
one tensor resident under the same budget prevents the failure. The two simple score changes we
tested do not remove holes across traces; both add failing budgets on some traces. All results are from simulation,
with no allocator model; behaviour on GPU hardware is untested.
\end{abstract}

\section{Introduction}
Rematerialization trades compute for memory: tensors are discarded during training and
recomputed when needed. Static planners choose what to recompute ahead of time
\cite{chen2016,jain2020checkmate}. DTR \cite{kirisame2021dtr} makes the choice online: when
an allocation does not fit the budget, it evicts the resident tensor with the lowest score
\begin{equation}
h_{\mathrm{DTR}}(t) = \frac{c(e^*(t))}{m(t)\cdot s(t)},
\end{equation}
where $c(e^*(t))$ is the compute of $t$ and of its evicted neighbourhood, $m(t)$ its size
and $s(t)$ its staleness. DTR is evaluated, like most rematerialization systems, by sweeping
the budget on a coarse grid and reporting overhead and the smallest budget that works
\cite{kirisame2021dtr,zhang2023coop,wang2026tcontrol}. Such coarse sweeps can fail to
reveal non-monotone feasibility; our results show that this is not merely hypothetical.

This paper reports that, in DTR's reference simulator, feasibility is not a threshold. At
some budgets DTR runs out of memory while slightly smaller and slightly larger budgets
complete. A feasibility hole is not an ordinary low-memory failure; it is a local failure
inside an otherwise feasible region. On InceptionV4, for example, DTR completes at a
budget of 0.2523 of peak memory and runs out of memory at 0.2527, and completes again at 0.26,
each outcome confirmed by repeated runs.

Online eviction policies are known to violate monotonicity in other settings. FIFO paging
can incur more faults with more memory (Belady's anomaly \cite{belady1969}). Stack
algorithms such as LRU cannot, because of the inclusion property \cite{mattson1970}. Greedy
schedulers show similar timing anomalies \cite{graham1969,reineke2006}. DTR's cost-based
score has no inclusion property, so non-monotone behaviour is possible in principle. The
question is whether it occurs on real training traces, how often a standard evaluation
would see it, and how it arises.

\paragraph{Contributions.}
\begin{enumerate}
\item Reproducible counterexamples to monotone feasibility for DTR in simrd: holes on two of
six confirmatory traces and on the development trace, each confirmed by repeats and by
unmodified simrd (Section~\ref{sec:holes}).
\item Evidence that a coarse sweep does not reveal them: the 0.01-step grid on the
confirmatory traces shows no DTR hole, although its unresolved timeouts prevent a conclusion
of monotonicity (Section~\ref{sec:holes}).
\item A step-by-step account of one failure, down to individual allocations, with an
intervention that prevents it under the same budget (Section~\ref{sec:mechanism}).
\item Negative results for two simple score changes, evaluated against criteria fixed
before the confirmatory runs (Section~\ref{sec:policies}).
\end{enumerate}
We make no claim about real GPU runtimes; Section~\ref{sec:limits} sets out what remains
untested.

\section{Background and related work}
\paragraph{Rematerialization.} Checkpointing in the style of Chen et al.\ \cite{chen2016}
and optimal static planning with Checkmate \cite{jain2020checkmate} decide recomputation
before execution. DTR \cite{kirisame2021dtr} is the online, greedy alternative and
introduced simrd, the simulator we use. It also proves an adversarial lower bound for any
deterministic heuristic, and notes that its heuristic can cause deeply nested
rematerializations. Later systems change the score or the memory manager. Coop
\cite{zhang2023coop} drops the size term and searches for contiguous evictions.
MegTaiChi/DTE \cite{hu2022megtaichi} adds fragmentation awareness. T-Control
\cite{wang2026tcontrol} locks high-betweenness ``hub'' tensors and evicts by segment,
explicitly to avoid deep recursive recomputation. Coop and T-Control evaluate budgets in steps of about 10\%, and these evaluations report
time per iteration or minimum budget. In the papers we could read,
none reports a budget that fails between two that succeed. The DTR paper reports that random sampling ``caused occasional
failures'' at some budgets in its PyTorch prototype, and that even without sampling DTR
``still occasionally failed on UNet'', possibly because of allocator details or variance in
operator times \cite{kirisame2021dtr}. A PyTorch issue reports peak memory that is non-monotone in
the compile-time activation-memory budget, attributed to recomputation chains
\cite{pytorch197838}; that concerns peak memory, not feasibility.

\paragraph{Prior search.} Our literature search (Appendix~\ref{app:lit}) did not identify a
prior characterisation of feasibility holes in DTR. Its coverage of papers citing DTR is
incomplete, so we make no priority claim.

\paragraph{This author's earlier preprint} \cite{pagadala2026regime} reported, on the
ResNet-32 trace, the failure band that serves here as the development case. It also
reported a regime switch in LSTM overhead. This paper tests whether those observations
generalise, under a written protocol on traces not used during development.

\section{Method}\label{sec:method}
\paragraph{Simulator, traces, budgets.} All runs use simrd at commit
\texttt{eff53cc4}. For the cloud runs this is a recorded git checkout. The laptop's copy
has no git metadata, and one batch's manifest recorded the wrong commit. A content check
showed that all 19 simulator source files and all trace files on the laptop are
byte-identical to that commit, with modification times that predate the laptop runs
(\texttt{SIMRD\_PROVENANCE.md}). This is a retrospective check: timestamps support, but do
not prove, that the same files were used at run time. The runs use the \texttt{RuntimeV2Eager\-Optimized} runtime and the public traces
released with DTR. simrd models tensor sizes, operator compute costs, eviction and
rematerialization; it has no allocator or fragmentation model. ResNet-32 is the development
trace: all exploratory tuning used it. The six confirmatory traces are DenseNet,
InceptionV4, Transformer, U-Net, TreeLSTM and an unrolled GAN. An LSTM trace is
reported separately. A budget is a fraction $r$ of the trace's unconstrained peak memory,
$\lfloor r \cdot \mathrm{peak}\rfloor$ bytes, the same form simrd's own evaluation scripts
use.

\paragraph{Instrumentation.} Our runtime subclasses simrd's runtime without changing any
decision. At every allocation check it records \emph{pinned} bytes (resident bytes outside
the evictable pool) and the rematerialization nesting depth. We report two depth measures
and never merge them: \emph{depth at peak pinned memory} and \emph{maximum nesting depth}.
Key points are rerun with \emph{unmodified simrd}: its runtime and DTR heuristic straight
from the commit, with no code of ours on the simulation path.

\paragraph{Outcomes and overhead.} A run ends \emph{ok}; \emph{OOM}, when an allocation
cannot be satisfied after the evictable pool is empty; \emph{thrashed}, when overhead
exceeds 60$\times$ (rematerialization compute above 59$\times$ the model's compute); \emph{recursion}, when the
call stack is exhausted; or \emph{timeout}, after 20 minutes of wall time. Only OOM is
evidence of memory infeasibility. Timeouts are unresolved and are never counted as
successes or failures. Compute overhead is (model compute + rematerialization compute) /
model compute, for completed runs. Every protocol run is a separate process.

\paragraph{Protocol.} The sweep was fixed in writing before its results existed, and changed
only through dated amendments, all in the repository.
\begin{itemize}
\item \emph{Stage A} runs DTR and two variants on a 0.01 grid from 0.05 to 0.60 on every
confirmatory trace, with three repeats at a seeded 10\% of budgets.
\item \emph{Stage C} adds seeded interior samples in every 0.01 interval (two below 0.30,
one above). It was added before any Stage A result existed on traces other than DenseNet,
because a grid cannot see a failure region lying inside one interval.
\item \emph{Stage B} refines at 0.001 steps every interval where neighbouring samples
change status or overhead by more than 25\%.
\item \emph{Confirmation stages} were each specified before they ran, after earlier stages
had shown the region. They run three repeats of DTR and one run of unmodified simrd at
selected points. A point is confirmed if every completed repeat agrees and unmodified simrd
gives the same outcome; a timeout is reported as unresolved. The first such stage did not
state how to treat a timeout. The second (Stage E) treated it as unresolved and confirmed a
point when all completed repeats agree; we apply that rule retrospectively to the first,
noting where it matters. The last batch used a stricter rule (any timeout prevents
confirmation), which we report under its own rule.
\end{itemize}
Interrupted runs (host sleep, a killed process, or an accidental second launcher) are kept
in the evidence table with their exclusion reason and rerun. The table has 6{,}243
executions, of which 5{,}583 are valid outcomes.

\paragraph{Counting holes.} Because budgets are sampled, we report three quantities:
\emph{failing samples} above a policy's first success; \emph{sample-supported failure
regions}; and \emph{edge brackets}. Timed-out samples are set aside first. A failure region
is then a maximal run of consecutive failing samples, in budget order, with a completing
sample immediately below and immediately above it. This matches the definition of a hole:
success on both sides. Timeouts lying between failing samples therefore do not split a
region, and timeouts between a failure and a success are reported but neither extend nor
close it. Edge brackets give the last completing and first failing sample at each end. We do
not report hole widths, and a run of failing samples does not show that every budget
between them fails. Every failing sample above a policy's first success has a completing
sample somewhere above it, because the top of every grid completes.

\section{Feasibility holes exist, and a coarse grid misses them}\label{sec:holes}

\paragraph{Determinism.} Across 5{,}583 valid executions, run on a Linux cloud container and
a Windows laptop (hosts inferred from the run logs), no budget has two completed runs that
disagree on outcome. Repeats also agree exactly on
overhead, pinned bytes and both depth measures. Unmodified simrd gives the same outcome as
our runtime wherever both completed.

\paragraph{The coarse grid.} On Stage A's 0.01 grid, DTR has \emph{no} failing budget above
its first success on any of the six confirmatory traces. The coarse sweep reveals no
inversion, but its 13 timed-out DTR grid points (on InceptionV4, TreeLSTM and Unrolled GAN)
mean it could not have established monotonicity either. It simply gives no sign of the holes
found below.

\paragraph{Finer sampling.} With Stages C and B, DTR shows sample-supported failure regions
on three confirmatory traces (Table~\ref{tab:regions}). DenseNet, Transformer and
TreeLSTM show none at the sampled resolution. TreeLSTM has 46 timed-out samples, 13 of them above DTR's first
success, so its absence of holes is weaker evidence.

\begin{table}[t]
\centering\small
\caption{DTR failure regions supported by samples (Stages A, B and C plus confirmation runs).
Brackets give the last completing and first failing sample on each side. ``Confirmed'' means
three agreeing repeats plus agreeing unmodified simrd at the listed points, except where noted:
0.2423 has two completed, agreeing repeats and one timeout (see note $^a$); the region remains
supported by the confirmed point at 0.235.}
\label{tab:regions}
\begin{tabularx}{\linewidth}{@{}l>{\raggedright\arraybackslash}X>{\raggedright\arraybackslash}Xl@{}}
\toprule
Trace & Edge brackets: ok $\mid$ OOM \dots\ OOM $\mid$ ok & Confirmed points & Status\\
\midrule
ResNet-32 (dev.) & 0.101 $\mid$ 0.102 \dots\ 0.106 $\mid$ 0.107 & all, 3 repeats; stock at 0.101/0.104/0.107 & confirmed\\
InceptionV4 & 0.209 $\mid$ 0.212 \dots\ 0.222 $\mid$ 0.223 & 0.209, 0.212, 0.222, 0.223 & confirmed\\
 & 0.2343 $\mid$ 0.235 \dots\ 0.2425 $\mid$ 0.243 & 0.2343, 0.235, 0.2423$^a$, 0.248 & confirmed\\
 & 0.2524 $\mid$ 0.2526 \dots\ 0.2527 $\mid$ 0.254 & 0.2523, 0.2527, 0.26 & confirmed\\
U-Net & 0.422 $\mid$ 0.423 \dots\ 0.425 $\mid$ 0.426 & 0.422, 0.424, 0.426 & confirmed\\
 & 0.431 $\mid$ 0.432 \dots\ 0.435 $\mid$ 0.436 & 0.431, 0.433, 0.436 & confirmed\\
Unrolled GAN & 0.236 $\mid$ 0.237 \dots\ 0.238 $\mid$ 0.239 & ok at 0.236, 0.239 only & not confirmed$^b$\\
\bottomrule
\end{tabularx}

\smallskip
\footnotesize $^a$ Two completed repeats OOM, one timed out; confirmed only under Stage E's
rule (timeouts unresolved, all completed repeats agree), applied retrospectively. The region is also confirmed at 0.235 with all repeats complete.
$^b$ Two OOM repeats at 0.237; the remaining runs at 0.237 and 0.238 timed out or were
interrupted. Single Stage B runs give OOM at both budgets.
\end{table}

\begin{figure}[t]
\centering
\includegraphics[width=\linewidth]{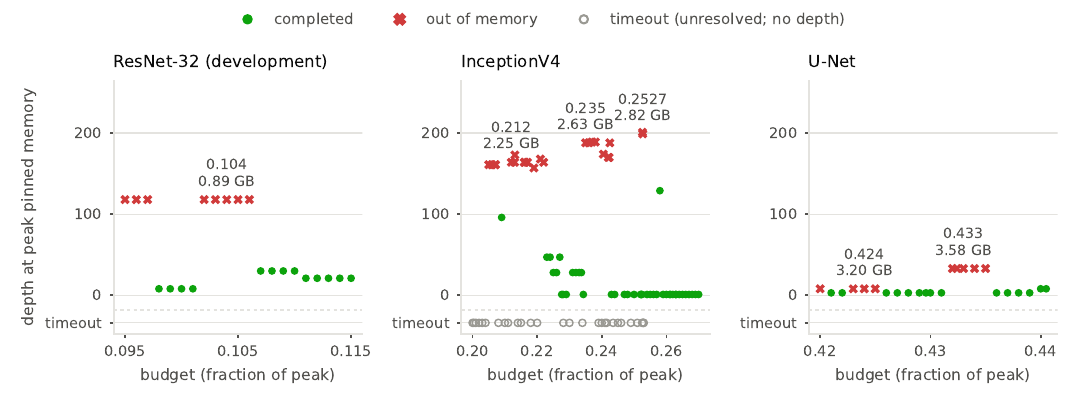}
\caption{DTR outcome against budget, with recursion depth at peak pinned memory on the
vertical axis. Green: completed; red: OOM. Hollow markers in the separate bottom strip are
timeouts (unresolved); their vertical position carries no depth. Annotated failures
are confirmed by three repeats and unmodified simrd.}
\label{fig:holes}
\end{figure}

Figure~\ref{fig:holes} shows the three confirmed traces. Three of the six confirmed holes
(ResNet-32 0.102--0.106, InceptionV4 0.2526--0.2527, U-Net 0.432--0.435) lie strictly inside
a 0.01 interval whose end points both complete. Stage A missed the other three for other
reasons. InceptionV4's two lower regions span grid points where Stage A timed out
(0.19--0.24). U-Net's lower region lies just above a failing grid point (0.42), below Stage
A's first success at 0.43. On
InceptionV4, DTR's feasibility switches six times between 0.209 and 0.26. Some samples
inside the failure regions (0.214, 0.215, 0.218, 0.22, 0.239--0.2415) and in the edge
brackets (0.2525, 0.2528--0.2530) timed out, so their outcome is unknown.

\paragraph{Failure signature.} On InceptionV4, every confirmed OOM reaches 2.25--2.82~GB of
pinned memory, with depth at peak pinned memory between 164 and 199. Confirmed successes in
the same range stay at 0.99--1.93~GB and depth 1--96. The signature is a strong
association, not a classifier. A completing run at 0.258 (11.2$\times$ overhead) reaches
2.67~GB at depth 129. On U-Net, the upper failure reaches depth 33 at peak pinned memory
(maximum nesting 36), but the lower one fails at depth 8 (maximum nesting 17) with 3.20~GB
pinned, so not every hole shows a deep cascade. The unconfirmed Unrolled GAN failures peak
at only 0.06~GB pinned, at depth 185 (maximum nesting 186).

\section{A worked example: how more memory produces a failure}\label{sec:mechanism}
We trace the InceptionV4 pair 0.2343 (completes, 1.471$\times$ overhead) and 0.235 (OOM).
The budgets are 2{,}634{,}785{,}825 and 2{,}642{,}657{,}571 bytes, 7{,}871{,}746 bytes apart.
Both outcomes are confirmed. The analyses in this section were specified before they ran,
after the earlier results were seen, and are exploratory.

\paragraph{Where the runs separate.} The two executions evict the same first 13 tensors and
first choose different victims at the 14th eviction (operator 945 of about 2{,}300). From
there to operator 2{,}269 their histories differ, but their event counts are close:
1{,}657 evictions each and 335 versus 334 rematerializations. Per-operator peak pinned
memory differs by less than 0.25~GB until operator 2{,}270 and is identical over operators
2{,}265--2{,}269. All of the failing run's extra work, 7{,}342 evictions and 6{,}143
rematerializations, happens inside one operator, 2{,}270.

\paragraph{Allocation threshold (operator 2{,}266).} Operator 2{,}266 produces storage 3131
(247{,}775{,}232 bytes). To make room, both runs make the same 14 evictions in the same
order. The smaller budget is then still 173{,}479 bytes short, so DTR evicts a fifteenth
storage, 932 (218{,}275{,}840 bytes). The larger budget is no longer short and keeps 932.

\paragraph{Changed victim (operator 2{,}267).} Operator 2{,}267 requests 60{,}211{,}200
bytes. At 0.2343 this fits in the space freed by storage 932. At 0.235 the run is
49{,}367{,}205 bytes short. Among 675 candidates, the lowest score belongs to storage 3131
($h = 2.82\times10^{-9}$). Second is storage 932 ($8.18\times10^{-9}$), the tensor the
smaller budget had evicted one operator earlier. DTR evicts 3131.

\paragraph{Cascade (operator 2{,}270).} Operator 2{,}270 reads storage 3131. At 0.2343 it is
resident and the run completes. At 0.235 it must be rebuilt, which recursively rebuilds its
own evicted ancestors. Maximum nesting depth reaches 193 and pinned memory peaks at 2.63~GB
(depth at peak pinned memory 188) before the evictable pool empties. That 3131 is needed
three operators after its eviction is hindsight from the trace. DTR's score does not use
future-access information.

\paragraph{Intervention.} We rerun 0.235 with one change: storage 3131 is never offered to
DTR as an eviction candidate, so its bytes stay resident within the same budget. The run
completes at 1.477$\times$ overhead, with 0.99~GB peak pinned memory and maximum nesting
depth 30. A control using the same code path and no retained tensor reproduces the original
failure exactly. Preventing storage 3131's eviction prevents OOM in this execution under the
same budget, which supports its causal involvement. Retaining it also changes the rest of
the execution, so this does not isolate it as the sole cause. Because the tensor is chosen
with hindsight, this intervention is not a policy.

\begin{figure}[t]
\centering
\includegraphics[width=\linewidth]{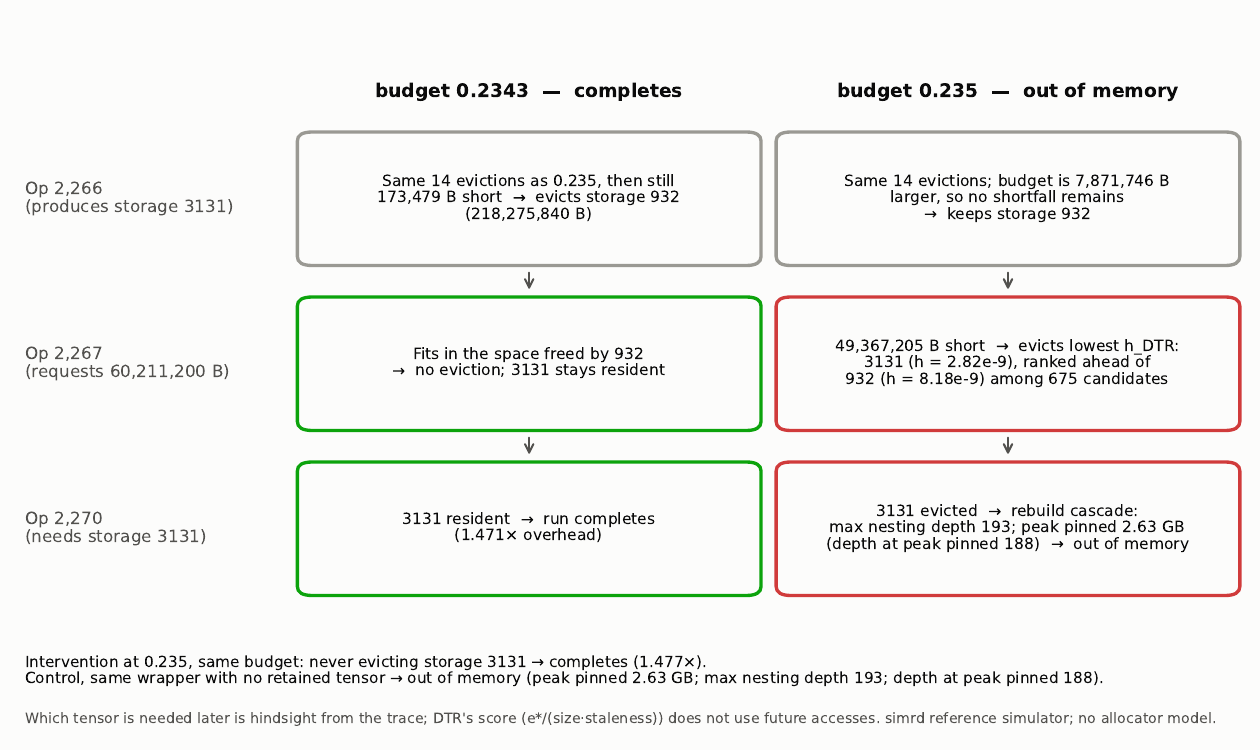}
\caption{The InceptionV4 pair at operators 2{,}266, 2{,}267 and 2{,}270. Exact values for
every decision are in the repository (\texttt{MECHANISM\_inception\_0.2343\_vs\_0.235.md}).}
\label{fig:mech}
\end{figure}

\paragraph{Summary.} In this execution, the extra 7.9~MB avoids one eviction at operator
2{,}266, so storage 932 (218~MB) stays resident going into operator 2{,}267. That forces an
eviction there, and DTR's cheapest choice is the tensor that operator 2{,}270 needs. More
memory did not give the same execution more room; it produced a different execution.

\paragraph{Supporting case.} On ResNet-32, the failing operator at 0.104 needs a storage
that the 0.104 run evicted five operators earlier and the 0.101 run never evicted.
Retaining it at 0.104 turns the failure into a completion (2.24$\times$), and the control
reproduces the failure. Our probe's memory accounting inside that operator is not yet
explained, so we report ResNet-32 only as a second instance of the retention result.

\section{Changing the score does not remove the holes}\label{sec:policies}
Two variants were fixed on the development trace before any confirmatory run.
\emph{NbhdPenalty} multiplies $h_{\mathrm{DTR}}$ by $(1 + 0.25\,|e^*|)$. \emph{TwoPhase}
keeps candidates large enough to cover the current shortfall (or, if none is, the eight
largest) and ranks them by cost over staleness. Against DTR, on each trace, the protocol tests: H1, no more failing samples
above the first success; H2, a matched-budget geometric-mean overhead ratio of at most
1.05; and H3, a stable-from budget no higher than DTR's. A policy's
\emph{stable-from} budget is the lowest sampled budget at and above which every completed
sample succeeds, with timeouts set aside. A timeout at or above it could only raise the true
value. H3 is therefore reported as unresolved only if treating such timeouts as failures
could change the verdict. A variant ``generalises'' only if H1 and H2 both hold on at least
five of six traces.

On ResNet-32, NbhdPenalty removes DTR's hole at every sampled budget. On the confirmatory
traces it does not generalise: H1 and H2 both hold on 2 of 6 traces (DenseNet,
Transformer). It has more failing samples than DTR on InceptionV4 (53 vs 16), U-Net (11 vs
8) and Unrolled GAN (7 vs 2), and a geometric-mean overhead ratio of 1.12 on TreeLSTM and
1.55 on Unrolled GAN. TwoPhase meets H1 and H2 on no trace. It has the most failing samples
(102 on InceptionV4) and failure regions on five traces, including DenseNet and
Transformer, where DTR has none. Every failure of the three frozen policies is an OOM.
Full tables are in the repository (\texttt{FINAL\_TABLES.md}).

Three further baselines were added before they ran, but only on the 0.01 grid: DTR's
$h_{e^*}$ variant, a Coop-inspired cost/staleness score, and a T-Control-inspired policy
that locks high-betweenness tensors. That grid shows no DTR hole on any trace, so it cannot
test whether a policy removes DTR's holes. The T-Control-inspired policy, which is our
simplified version and not the published system, has its own confirmed failure region on
InceptionV4 (ok at 0.20 and 0.23, OOM at 0.21 and 0.22, three repeats each). Among these
baselines, 24 runs stopped at the compute cap; we report them separately and do not count
them as holes.

\paragraph{Overhead is non-monotone too.} On the LSTM trace, DTR's overhead switches between
regimes: 10.10$\times$ at 0.266, 1.38$\times$ at 0.265, 8.78$\times$ at 0.264 and
1.39$\times$ at 0.262. They reproduce our earlier development runs on this trace exactly. NbhdPenalty shows the reverse pattern (1.31$\times$, 2.89$\times$, 1.31$\times$,
2.93$\times$), and the T-Control-inspired policy follows DTR's.

\section{Limitations}\label{sec:limits}
\begin{itemize}
\item \textbf{Simulation only.} simrd models no allocator, fragmentation, alignment,
streams or framework memory. A budget here is an exact limit on simulated tensor bytes. We
have not run DTR or any other runtime on hardware as part of this study, so whether these
feasibility holes transfer to hardware, and how often, is untested here. Fragmentation alone makes physical feasibility differ from the
logical budget in real DTR runtimes \cite{zhang2023coop,hu2022megtaichi}.
\item \textbf{Traces.} Each trace is a single recorded configuration from the DTR prototype.
Results say nothing about other models, batch sizes or frameworks.
\item \textbf{Sampling.} Holes are observed at samples; widths are not measured. Under
Stage C's sampling density and assuming holes are detected whenever a sample falls inside them,
a region of width 0.005 inside one interval has a 75\% detection probability below 0.30 and 50\%
above; these are detection probabilities under that assumption, not bounds on what remains
undiscovered.
\item \textbf{Unresolved runs.} 417 valid runs timed out. Some are probably failures, but we
do not assume so, and a timeout can also hide a run that would have hit the compute cap.
\item \textbf{Shared-defect risk.} All completed repeats agreed, and unmodified simrd
reproduced the selected outcomes. This is evidence that the outcomes were not introduced by
our instrumentation, not proof. It does not address a defect shared with simrd or the traces. Independent
reproduction is pending.
\item \textbf{Mechanism scope.} Section~\ref{sec:mechanism} explains one pair at the level
of allocations and supports a second through retention. We do not claim that other holes
arise by the same sequence; the U-Net lower hole, for example, shows no deep cascade.
\item \textbf{Protocol changes.} Stage C, the added baselines and every confirmation stage
were amendments. Each is dated, and results that depend on one are marked.
\end{itemize}

\section{Reproducibility}
Code, protocol, the dated amendment log and every execution record are at
\url{https://github.com/lonewolf15116/dtr-stability}. \texttt{results/EVIDENCE\_runs.csv}
lists all 6{,}243 executions, including interrupted and void runs with the reason they were
excluded. The ResNet-32 triple reproduces in seconds with a ten-line script that uses only
unmodified simrd (\texttt{REPRODUCE.md}). The InceptionV4 failing points take 13--20 minutes each.

\section{Conclusion and future work}
In DTR's reference simulator, feasibility is not monotone in the memory budget. Confirmed
holes occur on three of the seven traces we swept. On the confirmatory traces, the 0.01-step
sweep showed none of them, although its unresolved timeouts mean it could not establish
monotonicity either. In one
traced case the outcome turns on whether one allocation is 173{,}479 bytes short, through a
changed eviction choice that only matters three operators later. The two score changes we
tested did not remove the holes. Two questions remain open. The first is
whether holes occur in real runtimes: the original DTR prototype on GPUs, and modern
recomputation systems. In modern systems, budget settings select what to recompute rather
than enforcing a hard memory limit, so they need their own experimental design. The second
is whether a policy can guarantee monotone feasibility at acceptable cost, for example
through an inclusion-style property.

\paragraph{Acknowledgements and use of AI assistance.} AI assistants (Anthropic's Claude and
OpenAI's ChatGPT/Codex) were used extensively: to write and run experiment code, to audit
result files, to analyse data and to draft and fact-check this manuscript. Every number in
the paper is computed from execution records released in the repository. The author designed the study,
reviewed all results and the manuscript, and re-ran the ResNet-32 reproduction check.

\appendix
\section{Literature search}\label{app:lit}
We searched for reports of non-monotone feasibility in DTR and related systems, in the DTR
paper, Coop, T-Control (read in full), MegTaiChi, DELTA, Mimose, DyNN-Offload, Checkmate,
Moccasin, Rockmate, POET and others, plus targeted queries. The citation services we tried
were rate-limited and some publisher pages were blocked, so the list of DTR-citing papers is
a sample, not a census. The full log is in \texttt{LITERATURE\_SWEEP.md}.

\end{document}